\pdfoutput=1

\documentclass[11pt]{article}

\usepackage[preprint]{acl}

\usepackage{times}
\usepackage{latexsym}

\usepackage[T1]{fontenc}

\usepackage[utf8]{inputenc}

\usepackage{microtype}

\usepackage{inconsolata}

\usepackage{array}
\usepackage{color}
\usepackage{bm}
\usepackage{bbm}
\usepackage{enumitem}
\usepackage{booktabs}
\usepackage{amssymb}
\usepackage{amsfonts}
\usepackage{amsmath}
\usepackage{multirow}
\usepackage{graphicx}
\usepackage{amsfonts}
\usepackage[linesnumbered,ruled,vlined]{algorithm2e}
\usepackage{colortbl}
\usepackage{CJKutf8}
\usepackage{url}
\usepackage{here}
\usepackage{subcaption}
\usepackage{makecell}
\usepackage{mathrsfs}
\usepackage{titlesec}
\usepackage{lipsum}
\usepackage{booktabs}

\newcommand{\nsparagraph}[1]{\noindent\textbf{#1}\quad}
\newcommand{\subsectionmath}[1]{\subsection{\texorpdfstring{#1}{}}}

\title{Long-Tail Crisis in Nearest Neighbor Language Models}

\author{
  Yuto Nishida$^{1}$ $\;\;\;$ 
  Makoto Morishita$^{2}$ $\;\;\;$
  Hiroyuki Deguchi$^{1}$ $\;\;\;$ 
  \\
  \textbf{
  Hidetaka Kamigaito$^{1}$  $\;\;\;$ 
  Taro Watanabe$^{1}$
  }
  \\
  $^1$Nara Institute of Science and Technology $\;$ 
  $^2$Future Corporation $\;$ \\
  \texttt{\{nishida.yuto.nu8, deguchi.hiroyuki.db0, kamigaito.h, taro\}@is.naist.jp} \\
  \texttt{m.morishita.pi@future.co.jp}
}

\begin{document}
\maketitle
\begin{abstract}
The $k$-nearest-neighbor language model ($k$NN-LM), one of the retrieval-augmented language models, improves the perplexity for given text by directly accessing a large datastore built from any text data during inference.
A widely held hypothesis for the success of $k$NN-LM is that its explicit memory, i.e., the datastore, enhances predictions for long-tail phenomena.
However, prior works have primarily shown its ability to retrieve long-tail contexts, leaving the model's performance remain underexplored in estimating the probabilities of long-tail target tokens during inference.
In this paper, we investigate the behavior of $k$NN-LM on low-frequency tokens, examining prediction probability, retrieval accuracy, token distribution in the datastore, and approximation error of the product quantization.
Our experimental results reveal that $k$NN-LM does not improve prediction performance for low-frequency tokens but mainly benefits high-frequency tokens regardless of long-tail contexts in the datastore.\footnote{Our code and dataset are publicly available at \url{https://github.com/naist-nlp/knnlm-longtail-analysis}.}
\end{abstract}

\begin{figure*}[t]
    \centering
    \includegraphics[width=0.80\linewidth]{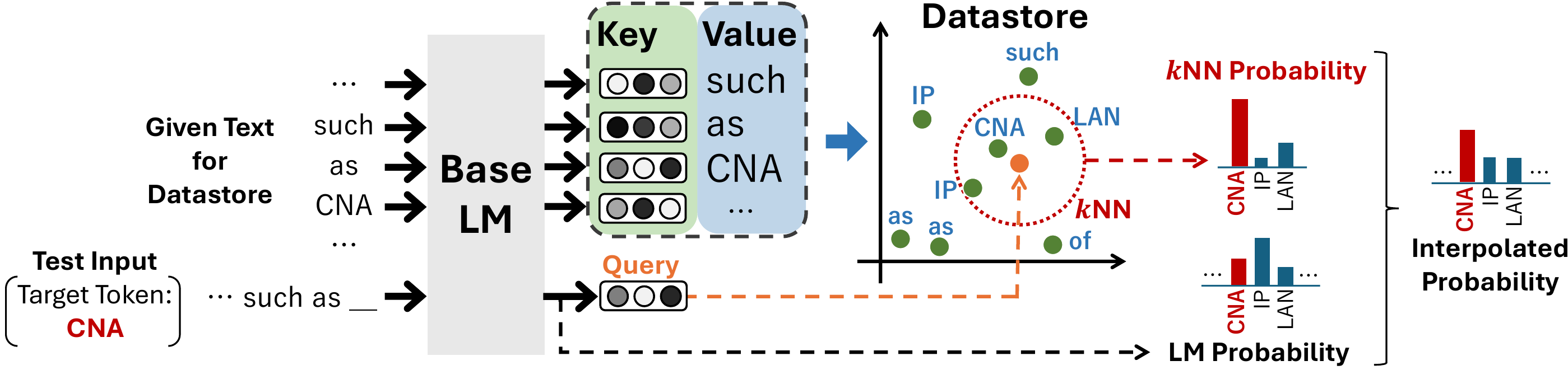}
    \caption{
    Overview of $k$NN-LM: The target token \textit{CNA} is retrieved as the nearest neighbor, and its prediction probability is enhanced by interpolating the $k$NN probability.
    \citet{Khandelwal2020Generalization} hypothesized that $k$NN-LM accurately predict long-tail phenomena, such as the low-frequency token \textit{CNA}, as shown in this example through the use of explicit memory (e.g., the datastore).
    }
    \label{fig:knnlm}
\end{figure*}

\section{Introduction}
\label{sec:introduction}
Retrieval-augmented language model~\cite{grave2017unbounded, guu-etal-2018-generating, Khandelwal2020Generalization, He2020Learning, borgeaud2022improving} is a novel paradigm for language modeling that incorporates a base language model (LM) with retrieved neighbor examples from an external datastore.
These models have been proven effective by impressive language modeling performance and achieved strong results on various tasks. 
One of the standard models in this paradigm is the $k$-nearest-neighbor language model ($k$NN-LM; \citealp{Khandelwal2020Generalization}), which interpolates the output probabilities of the trained base LM at each time step with $k$NN probabilities computed from retrieved neighbors in the datastore.
These neighbors are retrieved based on the distance between the contextualized embedding for each token in the test input and all key vectors in the datastore, which is constructed by caching intermediate representations for all tokens of external text data with the base LM.
Therefore, $k$NN-LM can accurately predict the probability of each token on a given test input by directly accessing text data through the datastore during inference.

As to why retrieval-augmented language models have been successful, several studies put forward an intriguing hypothesis that the explicit use of external memory in these models enhances prediction performance, particularly for low-frequency tokens~\cite{grave2017unbounded, merity2017pointer, li20ga_interspeech}.
In line with this, \citet{Khandelwal2020Generalization} argued that direct access to the datastore is beneficial for long-tail phenomena, providing several qualitative examples where $k$NN-LM successfully predicted target tokens even when given low-frequency contexts.
However, despite $k$NN-LM retrieving and incorporating the \emph{target token}, a token to predict, during inference, the prior work has focused primarily on long-tail \emph{context}, the sequence preceding the target token.
While it is expected that $k$NN-LM would improve predictions for long-tail target tokens as it does for long-tail contexts, our preliminary experiments revealed no correlation between the frequency of context $n$-grams and the frequency of target tokens (\S\ref{sec:preliminary_experiments}).
Thus, the behavior of $k$NN-LM with respect to long-tail target tokens remains unclear.

To further investigate this hypothesis, we conducted a detailed quantitative analysis, focusing specifically on long-tail target tokens.
We examined prediction probability, retrieval accuracy, and distribution of contextualized token embeddings in the datastore.

Our experiments on the resplit WikiText-103 dataset, using GPT2-XL as the base LM, revealed four key findings:
(1) The $k$NN probabilities for low-frequency tokens were lower than those of the base LM, whereas $k$NN probabilities for high-frequency tokens were higher (\S\ref{sec:knn_prob_analysis}). Thus, contrary to prior expectations, $k$NN-LM does not improve prediction performance for low-frequency tokens but instead enhances performance for high-frequency tokens.
(2) Nearest neighbor search is more challenging for low-frequency tokens, as most target low-frequency tokens were not included in the retrieved nearest neighbors (\S\ref{sec:knn_search_analysis}). This likely explains why $k$NN-LM struggles to predict low-frequency tokens.
(3) Low-frequency tokens have sparser distributions than high-frequency tokens in the datastore, and other tokens often appear mixed with the neighbors of low-frequency tokens (\S\ref{sec:datastore_distribution_analysis}). It complicates the nearest neighbor search.
(4) Long-tail tokens have larger reconstruction errors of the quantization in the datastore, making \textit{approximate} nearest neighbor search more challenging (\S\ref{sec:quantization_analysis}).

\section{Problem Statement}
\label{sec:problem_statement}
\subsectionmath{Preliminary: $k$NN-LM}
The $k$-nearest-neighbor language model ($k$NN-LM; \citealp{Khandelwal2020Generalization}) exploits knowledge of nearest neighbors retrieved from an external datastore with the base LM as shown in Figure~\ref{fig:knnlm}.

As a fundamental process in $k$NN-LM, we pre-construct a datastore using arbitrary text.
During inference, the $k$NN probabilities are calculated using the $k$-nearest-neighbor tokens retrieved from the datastore.
These probabilities are then interpolated with the base LM probabilities.
The $k$NN-LM improves the perplexity by utilizing explicit memory, i.e., the datastore.

\paragraph{Datastore Construction}
Let a sequence of tokens $\bm{x}=(x_{1},\cdots,x_{|\bm{x}|}) \in \mathcal{V}^*$ to be text stored in a datastore, where $\mathcal{V}$ denotes the vocabulary and $*$ represents the Kleene star.
The datastore $\mathcal{D} \subseteq \mathbb{R}^{D} \times \mathcal{V}$ is constructed as a set of key-value pairs for all tokens in the text:
\begin{equation}
\label{eq:datastore}
    \mathcal{D} =
    \left \{(\bm{f}_\theta(\bm{x}_{<t}), x_{t}) \right \}_{t=1}^{|\bm{x}|}.
\end{equation}
Each key is the intermediate representation $\bm{f}_\theta(\bm{x}_{<t})$ at each time step, where $\bm{f}_\theta \colon \mathcal{V}^* \to \mathbb{R}^{D}$ denotes the function that computes a $D$-dimensional hidden vector for the context $\bm{x}_{<t}$ using the base LM $\theta$. 
Each value is the target token $x_{t}$, i.e., the next token in each context. 
The key-value pairs are not stored by token type, i.e., vocabulary element, but encompass all tokens.
Consequently, the number of pairs in the datastore with a specific token type equals the frequency of that token type in the text from which the datastore is constructed.

The datastore often becomes too large because it stores high-dimensional contextualized representations for all tokens in the text.
Therefore, product quantization (PQ; \citealp{jegou2011product}) is commonly used in $k$NN-LM to compress the key vectors in the datastore.

\paragraph{$k$NN Probability Calculation}
When given a test input $\bm{x}$, the base LM $\theta$ computes a hidden vector $\bm{q}_{t}=\bm{f}_\theta(\bm{x}_{<t})$ for each target token $x_{t}$.
This vector is used as a search query to retrieve the $k$-nearest-neighbors $\mathcal{N}(\bm{q}_t; \mathcal{D}) \subset \mathcal{D}$ from the datastore $\mathcal{D}$.
The $k$-nearest-neighbors probability $p_{\text{$k$NN}}$ for the target token $x_{t}$ is then calculated based on the distance between the query $\bm{q}_{t}$ and the retrieved keys as follows:
\begin{align}
\label{eq:$k$NN_probability}
    &p_{\text{$k$NN}}(x_{t}|\bm{x}_{<t}) \nonumber \\
    &\propto 
    \sum_{(\bm{k},v) \in \mathcal{N}(\bm{q}_t; \mathcal{D})}{\mathbbm{1}_{x_{t}=v}\exp \left(\frac{-d(\bm{k},\bm{q}_{t})}{\tau}\right )}, 
\end{align}
where $d(\cdot,\cdot)$ is the distance function, and $\tau$ is the temperature parameter of the softmax function.\footnote{Note that the temperature parameter $\tau$ does not appear in the original $k$NN-LM~\cite{Khandelwal2020Generalization}. In this paper, we introduced the temperature to control the smoothness of the $k$NN probability distributions inspired by the $k$-nearest-neighbor machine translation ($k$NN-MT; \citealp{khandelwal2021nearest}), a variant of $k$NN-LM for encoder-decoder architectures.}
The $k$NN probability is assigned higher for tokens closer to the query than for those farther away.\footnote{The temperature parameter $\tau$ adjusts the influence of distance on probability assignment. A higher value of $\tau$ reduces the impact of distance on the assigned probabilities. In the limit, $\tau \rightarrow \infty$, the $k$NN probability for a token becomes independent of its distance and is equal to the relative count of that token in the neighbors.}
If a token is not retrieved in the neighbors, its $k$NN probability is assigned to 0.

\paragraph{Interpolation}
Given a context $\bm{x}_{<t}$, the prediction probability $p$ of the target token $x_{t}$ is computed by linear interpolation of the $k$NN probability $p_\text{$k$NN}$ and the prediction probability of the base LM $p_\text{LM}$ as follows:
\begin{align}
\label{eq:interpolation}
    &p(x_{t}|\bm{x}_{<t}) = \nonumber \\
    &\lambda p_\text{$k$NN}(x_{t}|\bm{x}_{<t})
    + (1 - \lambda)p_\text{LM}(x_{t}|\bm{x}_{<t}),
\end{align}
where $\lambda$ is a hyperparameter that controls the weights of the $k$NN probabilities.
Note that the prediction performance of $k$NN-LM on a target token improves when $p_\text{$k$NN}$ of the token is greater than $p_\text{LM}$ since the prediction probability $p$ is a weighted average of $p_\text{$k$NN}$ and $p_\text{LM}$, regardless of the value of the interpolation factor $\lambda$ when $\lambda > 0$.\footnote{Since $p = \lambda p_\text{$k$NN} + (1 - \lambda)p_\text{LM} > \lambda p_\text{LM} + (1 - \lambda)p_\text{LM} = p_\text{LM}$, i.e., the prediction probability $p$ is greater than the base LM probability $p_\text{LM}$ when $p_\text{$k$NN}>p_\text{LM}$ and $\lambda > 0$.}

During inference, $k$NN-LM directly accesses text data via the datastore.
\citet{Khandelwal2020Generalization} hypothesize that this explicit memory allows for better prediction of low-frequency phenomena, which is one factor in the success of $k$NN-LM.

\subsection{Research Questions}
Retrieval-augmented LMs, which explicitly use external memory, are hypothesized to be particularly effective for low-frequency tokens~\cite{grave2017unbounded,merity2017pointer,li20ga_interspeech}.
One such model, $k$NN-LM, is expected to improve predictions for low-frequency target tokens by interpolating $k$NN probabilities obtained from the datastore.
While prior work has demonstrated that $k$NN-LM improves predictions for low-frequency contexts, our preliminary experiments (\S\ref{sec:preliminary_experiments}) revealed no correlation between the frequency of context $n$-grams and the frequency of target tokens. 
Therefore, the behavior of $k$NN-LM for low-frequency target tokens remains unclear.
To address this gap, we conducted a detailed analysis of the relationship between token frequency and the behavior of the $k$NN-LM.
In particular, we investigate the following four research questions.

\nsparagraph{Research Question 1}
Does $k$NN-LM improve the prediction performance for low-frequency target tokens?
We investigate the relationship between token frequency and both the prediction probability of the base LM and the $k$NN probability.
We show that $k$NN-LM actually worsens the prediction of low-frequency target tokens but improves the prediction of high-frequency target tokens (\S\ref{sec:knn_prob_analysis}).

We then deeply probe the causes of $k$NN-LM's poor prediction performance for long-tail target tokens by conducting a detailed analysis of the $k$NN probability calculation process.

\nsparagraph{Research Question 2}
Does the retrieval module effectively search for low-frequency tokens?
Our analysis concludes that the failure to retrieve the majority of low-frequency tokens worsens their prediction with $k$NN-LM (\S\ref{sec:knn_search_analysis}).

We investigate the reason why retrieving long-tail tokens is challenging.

\nsparagraph{Research Question 3}
How does the distribution of tokens in the key vector space of the datastore depend on their frequency?
We show the distribution of the datastore makes it challenging to retrieve low-frequency tokens (\S\ref{sec:datastore_distribution_analysis}).

\nsparagraph{Research Question 4}
Are low-frequency tokens quantized accurately?
We show the \textit{approximate} nearest neighbor search with quantization tends to have large quantization errors on low-frequency tokens (\S\ref{sec:quantization_analysis}).

\section{Experimental Settings}
\label{sec:experimental_settings}
We describe the experimental setup for the analysis of $k$NN-LM, which is shared across all experiments and analyses throughout this work.\footnote{The detailed settings are described in Appendix~\ref{appendix:detailed_settings}.}

\subsection{Base LM}
We used GPT2-XL~\cite{radford2019language} as the base LM comprising 1.5B parameters.\footnote{We also conducted experiments with models of different sizes, namely GPT2-Medium (355M) and GPT2-Large (774M), and confirmed that the results remained consistent.}
This model was pre-trained on English web text data, which was crawled from outbound links on Reddit.
Note that GPT-2 intentionally excludes all Wikipedia pages from the pre-training data because Wikipedia is often used as test data in language modeling tasks.
It is popular to use GPT-2 as the base model for $k$NN-LM and is already proven to be effective in the previous research~\cite{shi-etal-2022-nearest,xu2023nearest,wang-etal-2023-knn}.

\subsection{Token Frequency}
We have investigated two types of frequencies when analyzing the low-frequency tokens in $k$NN-LM.
First, we analyzed the frequency of tokens in the texts used to construct the datastore for the $k$NN-LM (\emph{datastore frequency}).
Second, we have also analyzed the frequency of tokens in the texts used to pre-train the base LM (\emph{pre-training frequency}).
Since the training data for the base LM, GPT-2, is not publicly available, we used the token frequency from its replica, the OpenWebTextCorpus~\cite{Gokaslan2019OpenWeb}, as a proxy for the base LM's frequency.
By examining these frequencies, we aim to elucidate the relationship between token frequency and $k$NN-LM performance through comprehensive experiments.

\subsection{Dataset}
We used WikiText-103~\cite{merity2016pointer} for our evaluation.
The original split of WikiText-103 is unsuitable for analyzing the behavior of low-frequency tokens because of the large lexical overlap between test/valid data and train data.
Therefore, we resplit the original WikiText-103 training split into the new train, test, and valid data.
These test and valid data were split to contain many low-frequency tokens in the datastore constructed from the new training data.
The process involved:
\begin{enumerate}[topsep=2pt,parsep=2pt,itemsep=2pt]
    \item Tokenize the original training data using the subword tokenizer of the base LM.
    \item Extract documents based on the ratio of $n$-grams with a frequency of 1 to all $n$-grams in the original training data over $n=1,2,3,4$ until the number of tokens almost matches the original valid and test data combined.
    \item Divide the extracted documents randomly to create new valid and test datasets, then treat the remaining documents as training data.
\end{enumerate}

Table~\ref{tab:corpus_stats} shows the number of tokens of the original and resplit dataset.
We compared the token frequency distributions of the original and resplit datasets in Appendix~\ref{appendix:dataset}.
We confirmed that the resplit test data contains more low-frequency tokens than the original data, making it more suitable for our analysis conducted in this paper.

\begin{table}[t]
    \small
    \centering
    \begin{tabular}{@{}lrr@{}}
        \toprule
        Subset & Original & Resplit \\
        \midrule
        Train & 117,317,490 & 116,754,298 \\
        Validation & 245,757 & 276,476 \\
        Test & 280,573 & 289,787 \\
        \bottomrule
    \end{tabular}
    \caption{Dataset size: the number of tokens.}
    \label{tab:corpus_stats}
\end{table}

\subsectionmath{$k$NN-LM}
We used the \texttt{knn-transformers}~\cite{alon2022neuro} for the implementation of $k$NN-LM.
For datastore creation and nearest neighbor search, we employed \texttt{FAISS}~\cite{johnson2019billion,douze2024faiss}.
The squared-L2 was used as a distance function.
The keys in the datastore were 1,600-dimensional representations corresponding to the inputs of the feed-forward network in the final layer.
For efficiency, we quantized the datastore using IVFPQ~\cite{jegou2011product}.
Following \citet{Khandelwal2020Generalization}, the interpolation factor $\lambda$ was set to 0.25.
The datastores were constructed using both the original and resplit training data, and these were evaluated on the test data for the corresponding split.

\subsection{Preliminary Experiments}
\label{sec:preliminary_experiments}

\paragraph{Overall perplexity}
Before conducting a detailed performance analysis, we first measured the perplexity (PPL) of $k$NN-LM on both the original WikiText-103 and the resplit datasets.
The datastores for $k$NN-LM were constructed from the training data of each respective dataset, and the PPL was evaluated with the corresponding evaluation data.
The number of neighbors $k$ and the softmax temperature $\tau$ for $k$NN-LM were chosen to minimize the PPL on the validation data, resulting in $k=1024$ and $\tau=10$ for both the original and resplit data.

Table~\ref{tab:preliminary_ppl} compares PPL between the base LM and $k$NN-LM on the test data for each dataset.
The results show that $k$NN-LM outperforms the base LM on both the original and resplit datasets, indicating its effectiveness even with data containing many low-frequency tokens.
However, the improvement was less pronounced on the resplit data than the original data, suggesting that the increase of low-frequency tokens limits the effectiveness of $k$NN-LM.

In the following sections, we conduct a detailed analysis of the prediction of $k$NN-LM for each token in the resplit test data, where the overall PPL shows improvement.

\begin{table}[t]
    \small
    \centering
    \begin{tabular}{@{}lrr@{}}
        \toprule
        & \multicolumn{2}{c}{Perplexity $\downarrow$}\\
        \cmidrule{2-3}
        Model & Original & Resplit \\
        \midrule
        Base LM & 14.37 & 17.30 \\
        $k$NN-LM & \textbf{10.74} & \textbf{15.68} \\
        \bottomrule
    \end{tabular}
    \caption{PPL on the test data of WikiText-103.}
    \label{tab:preliminary_ppl}
\end{table}

\begin{figure}[t]
    \centering
    \includegraphics[width=0.9\linewidth]{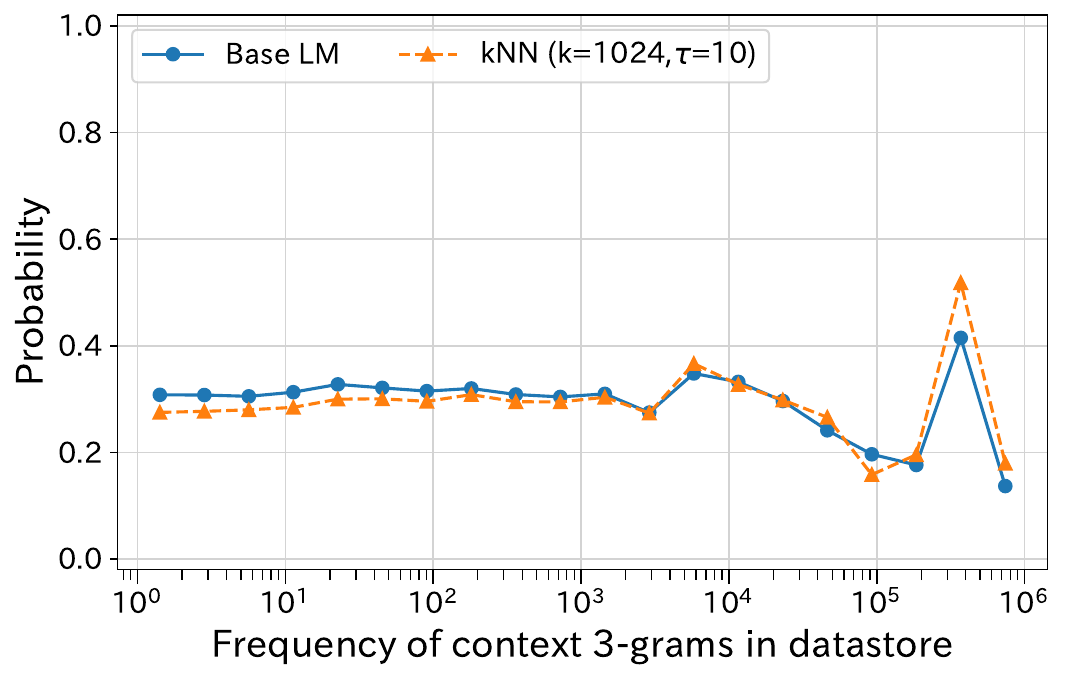}
    \caption{The relationship between the frequency of context 3-grams in the datastore and the expected values of $k$NN/LM probabilities on the resplit test.}
    \label{fig:ngram_prob}
\end{figure}

\paragraph{Predictions for long-tail contexts}
Retrieval-augmented LMs, such as $k$NN-LM, are hypothesized to be particularly effective for long-tail phenomena~\cite{grave2017unbounded,merity2017pointer,li20ga_interspeech}.
\citet{Khandelwal2020Generalization} provided qualitative examples showing that $k$NN-LM improves predictions for low-frequency context, i.e., the sequence preceding the target token.
However, when we examined the relationship between the frequency of context $n$-grams and prediction probabilities using the resplit test, we found that, as shown in Figure~\ref{fig:ngram_prob} for $n=3$, both $k$NN and LM probabilities exhibit similar trends, regardless of the frequency of the context.\footnote{We showed the results for context $n$-grams with other $n$ values ($n=1, 2, 4, 5$) in Appendix~\ref{appendix:context}.}
This indicates that $k$NN-LM does not particularly improve predictions for long-tail contexts.
Moreover, since the datastore of $k$NN-LM is constructed as pairs of keys, i.e., the intermediate representations of the context, and values, i.e., the target tokens, long-tail phenomena are primarily divided into long-tail contexts and long-tail target tokens.

Therefore, to fully understand $k$NN-LM's behavior on long-tail phenomena, we should analyze $k$NN-LM focusing on the target tokens.
To investigate whether the relationship observed for contexts holds for target tokens as well, we show the correlation between the frequency of context $n$-grams ($n=1...5$) and the frequency of target tokens in Table~\ref{tab:preliminary_corr}.
Since there is no correlation between contexts and targets, we independently analyze the behavior of $k$NN-LM for long-tail target tokens, investigating the factors contributing to the success of $k$NN-LM.

\begin{table}[t]
    \small
    \centering
    \begin{tabular}{@{}ccccc@{}}
        \toprule
        $n=1$ & $n=2$ & $n=3$ & $n=4$ & $n=5$ \\
        \midrule
        $-0.12$ & $-0.067$ & $-0.050$ & $-0.012$ & $-0.019$ \\
        \bottomrule
    \end{tabular}
    \caption{Pearson correlation coefficient between frequency of target tokens and frequency of context $n$-grams ($n=1,\cdots,5$) on the test data of WikiText-103.}
    \label{tab:preliminary_corr}
\end{table}

\section{Analyses}
\subsectionmath{$k$NN-LM Does NOT Improve Prediction Probability for Long-Tail Tokens}
\label{sec:knn_prob_analysis}
\citet{Khandelwal2020Generalization} hypothesized that $k$NN-LM improves prediction performance for long-tail phenomena by utilizing explicit memory, namely, the datastore, and they demonstrated improvements in predicting long-tail contexts.
To further investigate this hypothesis, we focus on the target tokens, which are actually retrieved by $k$NN-LM, rather than the context.
We perform a quantitative analysis to examine the relationship between token frequency and prediction probability.
According to Equation~\ref{eq:interpolation}, $k$NN-LM's prediction improves when the $k$NN probability exceeds the output probability of the base LM.
Therefore, we compared these probabilities to determine whether the $k$NN probability surpasses the base LM's output probability for low-frequency tokens.

\begin{figure}[t]
    \centering
    \includegraphics[width=0.91\linewidth]{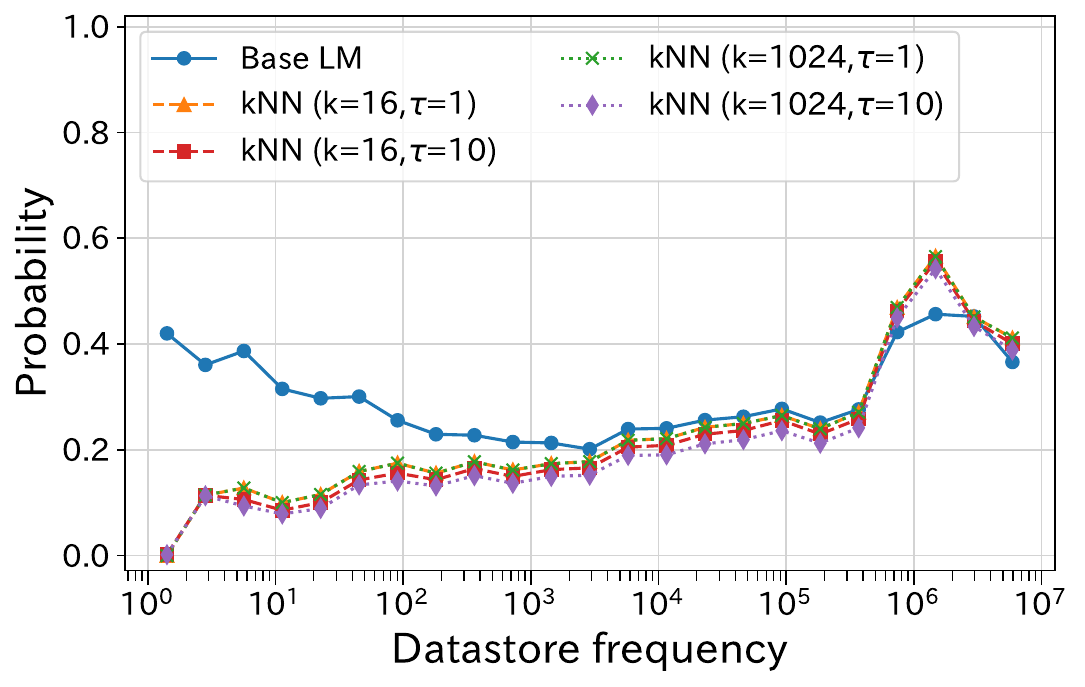}
    \caption{The relationship between datastore frequency and the expected values of $k$NN/LM probabilities on the resplit test: 
    At low frequencies, the $k$NN probability was lower than the LM probability.
    At high frequencies, the opposite trend was observed.}
    \label{fig:knn_prob}
\end{figure}

Figure~\ref{fig:knn_prob} illustrates the relationship between the token frequency of the datastore, and the expected values of the $k$NN probability and the base LM probability on the resplit test data.
The figure shows that for tokens with a frequency of approximately $10^5$ or less in the datastore, the $k$NN probability was lower than the base LM probability, with the difference being particularly pronounced for frequencies of $10^3$ or less.
This result indicates that interpolating $k$NN probabilities with LM probabilities worsens prediction performance for low-frequency tokens.
Conversely, the expected value of the $k$NN probability for tokens with a frequency exceeding $10^5$ was higher than or comparable to the LM probability.

\begin{figure}[t]
    \centering
    \includegraphics[width=0.91\linewidth]{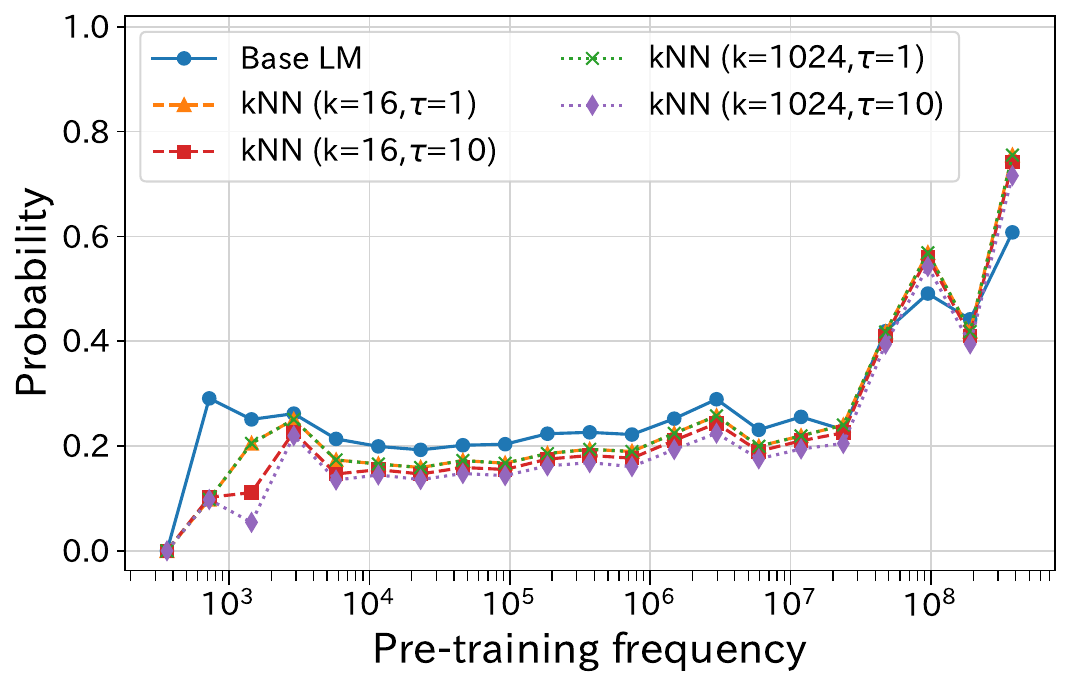}
    \caption{The relationship between pre-training frequency and the expected values of $k$NN/LM probabilities on the resplit test: 
    At low frequencies, the $k$NN probability was lower than the LM probability.
    At high frequencies, the opposite trend was observed.}
    \label{fig:knn_prob_openwebtext}
\end{figure}

Figure~\ref{fig:knn_prob_openwebtext} presents the relationship between the token frequency measured on the pre-training data for the base LM, and the expected values of probabilities on the resplit test data.
The figure reveals a similar pattern: at low frequencies, the $k$NN probability was lower than the LM probability, whereas at high frequencies, the opposite trend was observed.

For further analysis of low-frequency tokens, Appendix~\ref{appendix:knn_prob} compares the $k$NN probability and the base LM probability across different hyperparameters.
The $k$NN probability for low-frequency tokens was consistently lower than the LM probability, regardless of the number of neighbors $k$, and this disparity became more pronounced as the temperature $\tau$ increased.
Therefore, the tendency of $k$NN-LM to worsen prediction probabilities for low-frequency tokens, as we revealed, cannot be resolved by adjusting the hyperparameters.

To better understand the relationship between token frequency and prediction probability through case studies, Table~\ref{tab:example} shows qualitative examples in the resplit test data.
For target tokens \textit{intoxicated}, \textit{FreeBSD}, and \textit{recommendation} whose datastore frequencies were $10^3$ or less, the $k$NN probabilities were lower than the base LM probabilities.
These tokens, observed in sample contexts, would be related to the factual knowledge that \citet{Khandelwal2020Generalization} hypothesized $k$NN-LM would excel at; however, actually, $k$NN-LM worsens the predictions for them.
In contrast, the target token \textit{song}, whose frequency was approximately $10^5$, had a higher $k$NN probability than the base LM probability.

In summary, our findings demonstrate that \emph{$k$NN-LM does not improve the prediction performance of long-tail tokens} irrespective of the hyperparameters.
Instead, it suggests that $k$NN-LM enhances the prediction performance for high-frequency tokens.

\begin{table*}[t]
    \small
    \centering
    \tabcolsep 4pt
    \begin{tabular}{@{}lrcccp{6.7cm}p{1.5cm}@{}}
        \toprule
        & \multicolumn{1}{c}{\multirow{3}{*}{\makecell{Datastore \\ freq.}}} & \multicolumn{2}{c}{Probability $\uparrow$} & \multicolumn{1}{c}{\multirow{3}{*}{\makecell{$k$NN \\ hit rate $\uparrow$}}} & & \multicolumn{1}{c}{\multirow{3}{*}{\makecell{Top-3\\neighbors}}} \\
        \cmidrule(lr){3-4}
        \multicolumn{1}{c}{Target token} & & $k$NN & LM & & \multicolumn{1}{c}{A sample context for the target token of test input} & \\
        \midrule
        \verb*| intoxicated| & 5 & .137 & .518 & .600 & \textit{...They also put honey bees in shuttle-boxes that used a stimulus to encourage the bees to move, and found that they were less mobile as they became more \_\_\_} & \textit{in}, \textit{with}, \textit{intoxicated}\\
        \midrule[.03em]
        \verb*| FreeBSD| & 6 & .029 & .320 & .520 & \textit{...FreeBSD is a free Unix-like operating system descended from Research Unix via the Berkeley Software Distribution ( BSD ). Although for legal reasons \_\_\_} & \textit{the}, \textit{Perl}, \textit{a} \\
        \midrule[.03em]
        \verb*| recommendation| & 966 & .016 & .309 & .571 & \textit{...Sleep hygiene is the recommended behavioral and environmental practice that is intended to promote better quality sleep. This \_\_\_} & \textit{direct}, \textit{direct}, \textit{is} \\
        \midrule[.03em]
        \verb*| song| & 67,141 & .525 & .421 & .989 & \textit{...Yusuke Tanaka directed the accompanying music video for the single, which shows Perfume as robots and dancing with fairy lights around them. Perfume have performed the \_\_\_} & \textit{song}, \textit{song}, \textit{song} \\
        \bottomrule
    \end{tabular}
    \caption{Qualitative examples ($k=1024$, $\tau=10$):
    For target tokens\protect\footnotemark \textit{intoxicated}, \textit{FreeBSD}, and \textit{recommendation} whose datastore frequencies were $10^3$ or less, the $k$NN probability was lower than the base LM probability, and the $k$NN hit rate was below 60\%.
    In contrast, token \textit{song}, whose frequency was approximately $10^5$, had a higher $k$NN probability than the base LM probability, and in most cases, the nearest neighbors contained the target token.
    The right side of the table shows the sampled contexts for each token type in the resplit test data and displays the top-3 nearest neighbors for those contexts.
    For low-frequency tokens, the nearest neighbor for each context was an incorrect token, whereas for the high-frequency token, the correct target token was included in the nearest neighbors.}
    \label{tab:example}
\end{table*}

\subsectionmath{$k$NN Search for Long-Tail Tokens Is Challenging}
\label{sec:knn_search_analysis}
We demonstrated that the $k$NN probabilities are lower for long-tail tokens in \S\ref{sec:knn_prob_analysis}.
As a reason for this, we hypothesize that low-frequency tokens are less likely to be retrieved in the $k$NN search.
To validate this hypothesis, we analyzed the $k$NN hit rate, which is the proportion of the neighbors $\mathcal{N}$ that contain at least one instance of the target token.

\footnotetext{In GPT-2, tokenization is applied to the entire string, including spaces, so tokens sometimes contain spaces.}

\begin{figure}[t]
    \centering
    \includegraphics[width=0.9\linewidth]{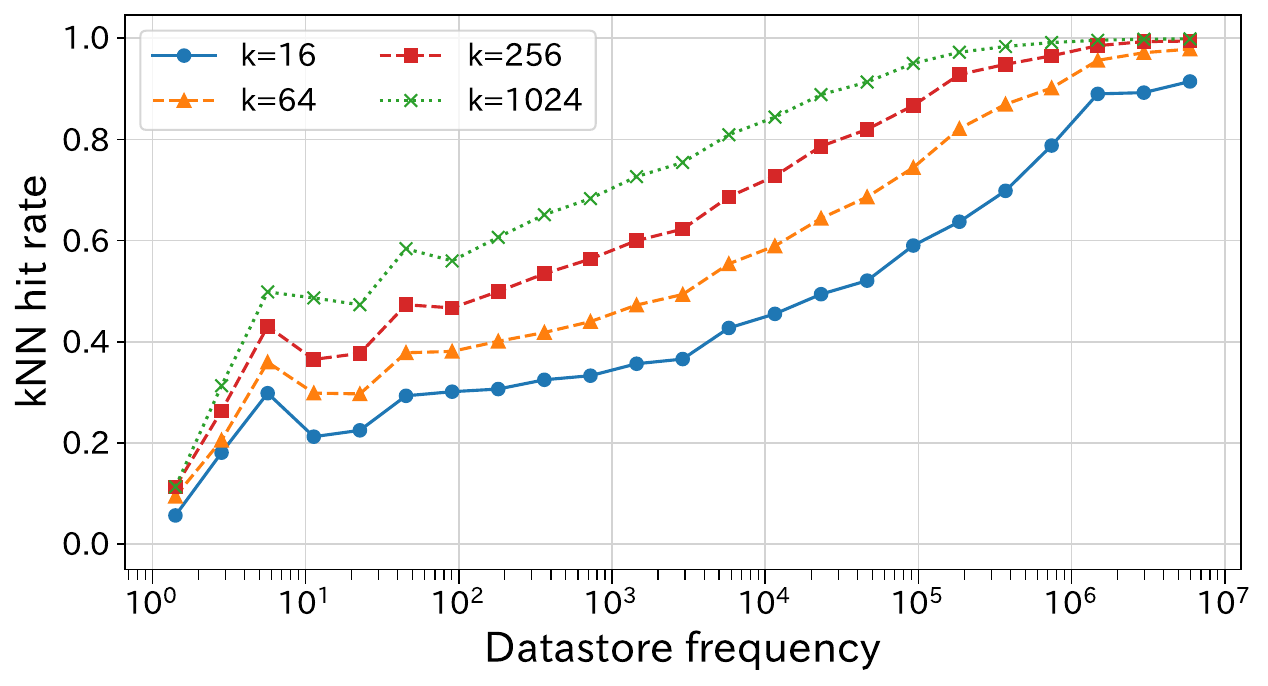}
    \caption{The relationship between datastore frequency and the expected $k$NN hit rate on the resplit test: For low-frequency tokens, the target token was not included in the neighbors at all.}
    \label{fig:knn_hit_ratio}
\end{figure}

Figure~\ref{fig:knn_hit_ratio} shows the relationship between the token frequency of the datastore and the expected $k$NN hit rate.\footnote{Figure~\ref{fig:knn_hit_ratio_openwebtext} shows the plot for token frequency of the pre-training data in Appendix~\ref{appendix:knn_hit_rate}.}
The figure indicates that increasing the number of neighbors $k$ tends to improve the $k$NN hit rate regardless of token frequency.
For tokens with frequencies of $10^5$ or more, the $k$NN hit rate approached nearly $100$\% with $k=1024$.
In contrast, for tokens with frequencies below $10^4$ with $k=16$, and below $10$ with $k=1024$, the $k$NN hit rate fell under 50\%.
This means that for most of the low-frequency tokens, \emph{the target token does not appear among the neighbors at all}, resulting in the $k$NN probability of 0.
This explains the low $k$NN probabilities for low-frequency tokens discussed in \S\ref{sec:knn_prob_analysis}.

Additionally, increasing the number of neighbors $k$ introduces more non-target tokens as noise, thereby lowering the relative frequency of instances with the value of the target token among the neighbors.
This poses a trade-off between the relative frequency of the target token and the $k$NN hit rate.
This trade-off would explain why the $k$NN probabilities do not improve with a larger number of neighbors, as observed in \S\ref{sec:knn_prob_analysis}.

We present qualitative examples of the $k$NN hit rate in the resplit test data in Table~\ref{tab:example}.
The table shows that target tokens \textit{intoxicated}, \textit{FreeBSD}, and \textit{recommendation}, which have low frequency in the datastore, had low $k$NN hit rates.
In contrast, the token \textit{song} with high frequency had a high hit rate.

\subsection{Long-Tail Tokens Are Distributed Sparsely with Contamination in Datastore}
\label{sec:datastore_distribution_analysis}
We showed that long-tail tokens were less likely to be retrieved during the $k$NN search in \S\ref{sec:knn_search_analysis}.
We hypothesize that this is due to the distribution of tokens in the datastore.
This section further analyzes the distribution in the datastore from two perspectives: the sparsity and contamination of neighbors for low-frequency tokens.

\paragraph{Sparsity}
Intuitively, if the keys of a target token type are sparsely distributed, the distances from these keys to the query corresponding to the target token type are expected to be larger, making the nearest neighbor search more challenging.
We hypothesize that low-frequency tokens are sparsely distributed in the datastore and investigate the relationship between token frequency and the dispersion of their distribution.

To quantify the sparsity of tokens with a token type, we measured the Euclidean distance from the centroid of the keys whose values correspond to the token type to each key.
We then calculated the coefficient of variation (CV) of these distances, i.e., the ratio of the standard deviation to the mean, which indicates the dispersion of the distribution.

\begin{figure}[t]
    \centering
    \includegraphics[width=0.9\linewidth]{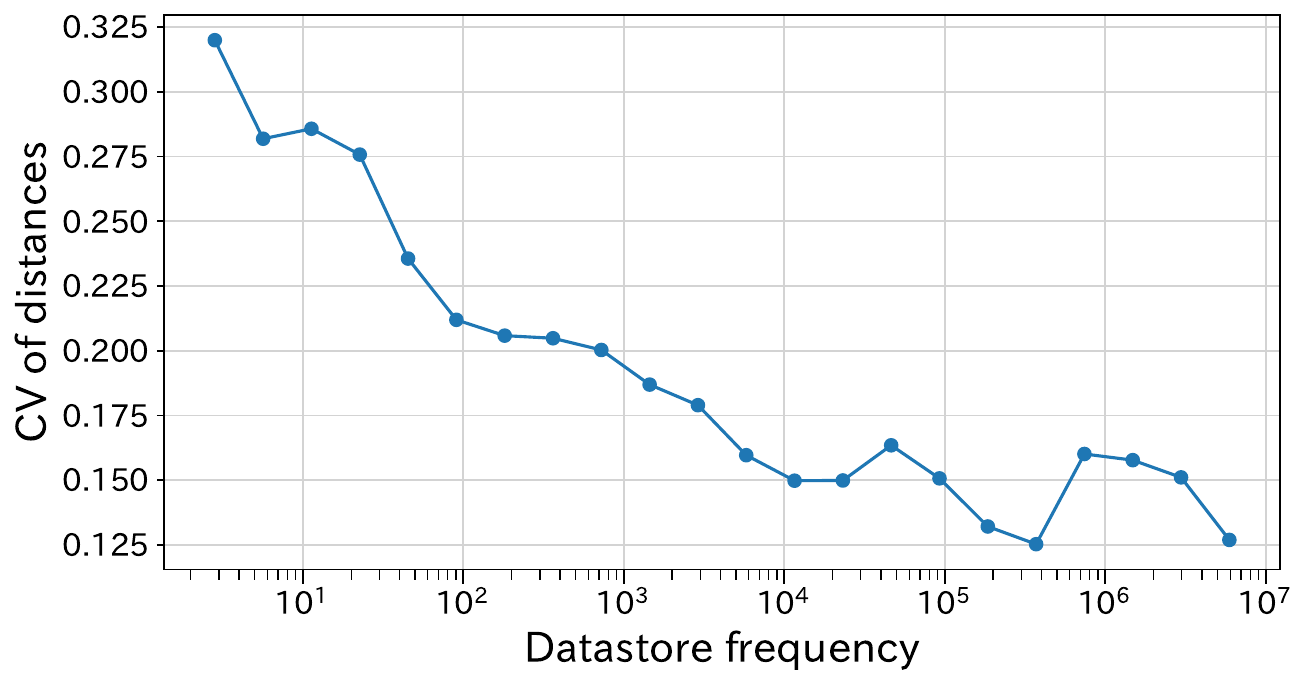}
    \caption{The relationship between token frequency of the datastore and the CV of the distances: Low-frequency tokens in the datastore have a higher CV compared to high-frequency tokens.}
    \label{fig:datastore_cv}
\end{figure}

Figure~\ref{fig:datastore_cv} shows the relationship between token frequency of the datastore and the CV of the distances.
The figure reveals that low-frequency tokens have a higher CV compared to high-frequency tokens.\footnote{Figure~\ref{fig:datastore_cv_openwebtext} shows the plot for token frequency of the pre-training data in Appendix~\ref{appendix:token_distribution}.}
This result implies that \emph{low-frequency tokens are more sparsely distributed in the datastore}, which suggests that their sparse distribution makes the search more challenging.

\paragraph{Contamination}
If the neighbors of the target token are contaminated with the non-target tokens in the datastore, retrieving the target token as the nearest neighbor becomes challenging.
We hypothesize that the neighbors of low-frequency tokens have different token types as the nearest neighbor, which might directly impact the $k$NN probabilities.

To validate this hypothesis, we investigated whether each instance in the datastore had the same token type as the nearest neighbor.
Specifically, we quantified the contamination rate for each token type.
We calculated the proportion of instances whose nearest neighbor instance had a different token type in the datastore.

\begin{figure}[t]
    \centering
    \includegraphics[width=0.9\linewidth]{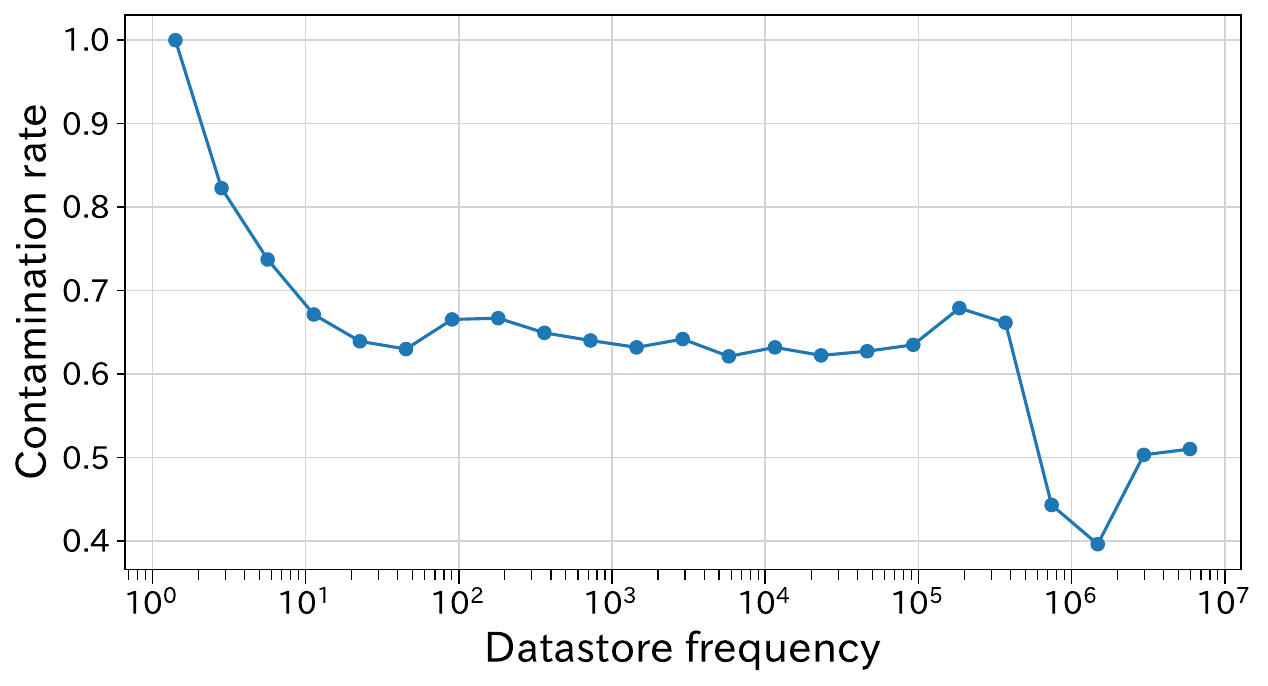}
    \caption{The relationship between the token frequency of the datastore and the contamination rate of the token type: Lower-frequency token types had higher contamination rates.}
    \label{fig:datastore_contamination}
\end{figure}

Figure~\ref{fig:datastore_contamination} shows the relationship between token frequency of the datastore and the contamination rate of the token type.\footnote{Figure~\ref{fig:datastore_contamination_openwebtext} shows the plot for token frequency of the pre-training data in Appendix~\ref{appendix:token_distribution}.}
The figure reveals that lower-frequency token types have higher contamination rates.
For token types with a frequency of less than 10, the nearest neighbor of the instance has a different token type in most cases.
This result indicates that \emph{low-frequency token types have contaminated neighbors in the datastore}, making the $k$NN search more difficult.

\begin{figure}[t]
    \centering
    \includegraphics[width=0.9\linewidth]{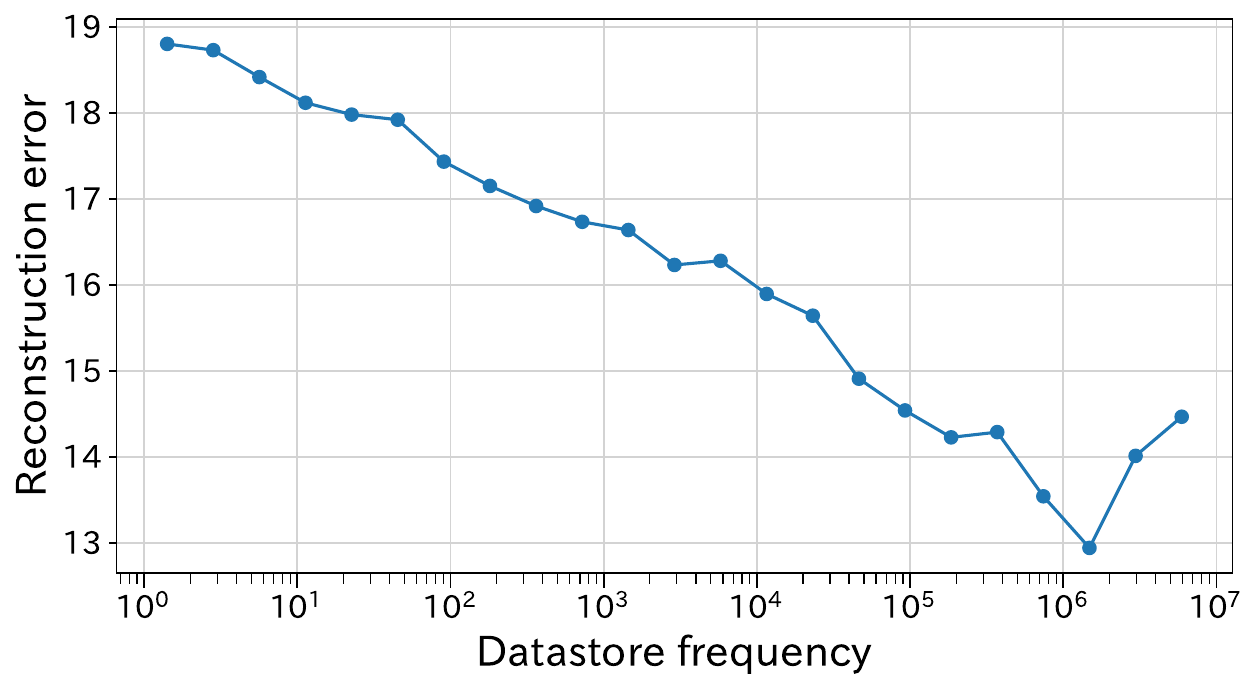}
    \caption{The relationship between the token frequency of the datastore and the reconstruction error of PQ: Low-frequency tokens have higher approximation errors in the datastore.}
    \label{fig:reconstruction_error}
\end{figure}

\subsection{Long-Tail Tokens Have Large Quantization Error in Datastore}
\label{sec:quantization_analysis}
We hypothesize another reason why the nearest neighbor search for low-frequency tokens fails is the adverse impact of approximation errors in the search process.
The product quantization (PQ; \citealp{jegou2011product}), commonly used for compressing key vectors in the datastore of $k$NN-LM, may accurately approximate the majority of tokens but might fail to approximate long-tail tokens.

We measured the approximation error introduced by datastore quantization as the average of the reconstruction error, defined as the Euclidean distance between the original key and the vector reconstructed from the PQ-encoded key.

\begin{figure}[t]
    \centering
    \includegraphics[width=0.9\linewidth]{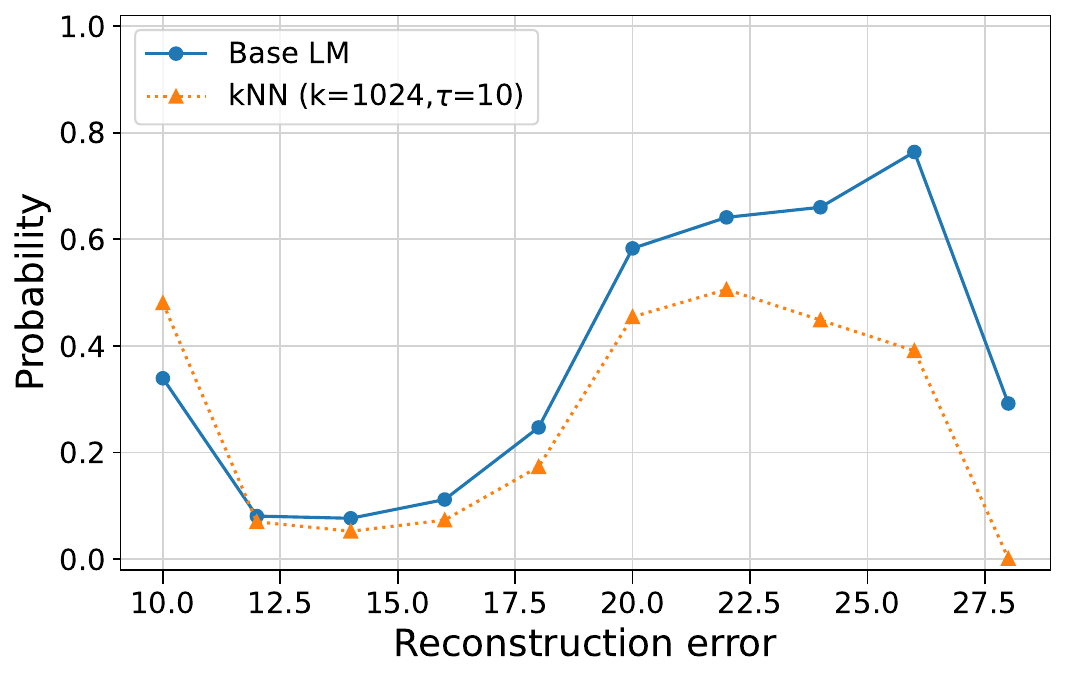}
    \caption{The relationship between $k$NN and base LM probability on the resplit test and the reconstruction error of the product quantization.}
    \label{fig:reconstruction_error_vs_probability}
\end{figure}

Figure~\ref{fig:reconstruction_error} shows the relationship between token frequency of the datastore and the reconstruction error of the PQ.\footnote{Figure~\ref{fig:reconstruction_error_openwebtext} shows the plot for token frequency of the pre-training data in Appendix~\ref{appendix:quantization_error}.}
The figure indicates that the quantization error is larger for low-frequency tokens and smaller for high-frequency tokens.
We then investigated the relationship between approximation error and $k$NN-LM's prediction performance.
Figure~\ref{fig:reconstruction_error_vs_probability} shows the relationship between $k$NN and base LM probabilities and PQ reconstruction error on the resplit test.
When the error was small, the $k$NN probabilities exceeded the base LM probabilities, leading to improved predictions through $k$NN augmentation.
In contrast, larger errors led to a decline in prediction performance.
Additionally, we calculated the correlation between the average prediction gain of $k$NN-LM (i.e., $p_\text{kNN} - p_\text{LM}$) and the average reconstruction error in the datastore for each token type.
The Pearson correlation coefficient was $-0.20$, and the Spearman's rank correlation coefficient was $-0.16$.
These negative correlations suggest that $k$NN-LM worsens the prediction performance for tokens with larger approximation errors in the datastore.
Thus, \emph{low-frequency tokens have higher approximation errors in the datastore}, making the approximate nearest neighbor search more difficult.

\section{Related Work}
Retrieval-augmented language model~\cite{grave2017unbounded,kaiser2017learning,guu-etal-2018-generating, Khandelwal2020Generalization, He2020Learning, borgeaud2022improving} incorporates a base language model (LM) with retrieved neighbor examples from a datastore for language modeling and other tasks.
\citet{kaiser2017learning} introduced a memory module capable of remembering rare events into a neural network and used it for text generation.
\citet{Khandelwal2020Generalization} proposed $k$NN-LM, one of the standards for retrieval-augmented LMs, which interpolates the output probabilities of the trained base LM, with $k$NN probabilities computed from the retrieved neighbors in the datastore.
Similar to \citet{kaiser2017learning}, they hypothesized that using an explicit memory, i.e., datastore, is particularly helpful for predicting long-tail phenomena based on qualitative examples.
While they demonstrated that $k$NN-LM improves the prediction of long-tail contexts, our detailed analysis reveals that $k$NN-LM often fails to enhance the prediction of long-tail target tokens, regardless of the frequency of the context $n$-grams.

To understand the strengths and limitations of $k$NN-LM, several studies have investigated the key factors underlying its performance.
\citet{drozdov-etal-2022-cant} find that large overlapping $n$-grams between the datastore and the evaluation data is crucial for the strong performance of $k$NN-LM.
This finding aligns with our results, showing that $k$NN-LM performs poorly on the resplit evaluation data of WikiText-103, which has less overlap with the training data.
\citet{wang-etal-2023-knn} revealed that focusing on individual tokens, the majority of tokens' predictions deteriorate with $k$NN-LM, leading to lower performance in open-ended text generation.
Our deep probing of individual tokens revealed that this majority primarily consists of low-frequency tokens.
\citet{xu2023nearest} implies through various ablation studies that memorization is not a key factor in the success of $k$NN-LM.
Our analysis of rare tokens further supports this hypothesis.

Several methods have been proposed to enhance the performance of $k$NN-LM and its derivative, $k$-nearest-neighbor machine translation ($k$NN-MT; \citealp{khandelwal2021nearest}).
These methods include adaptively determining the number of neighbors $k$ and the interpolation factor $\lambda$~\cite{zheng-etal-2021-adaptive,jiang-etal-2021-learning,jiang-etal-2022-towards}, as well as recalculating distances using two different datastores~\cite{bhardwaj-etal-2023-adaptation}.
However, these approaches do not explicitly address the negative impact on the prediction of low-frequency tokens, as demonstrated in this paper.

\section{Conclusion}
In this paper, we analyzed the behavior of $k$NN-LM, focusing on low-frequency target tokens.
Our analysis revealed that $k$NN-LM does not improve the prediction for long-tail target tokens, regardless of the frequency of the context.
This lack of improvement can be attributed to the challenges of $k$NN search for long-tail tokens, which is complicated by factors such as the distribution of the datastore and quantization errors.

Based on our findings, improving $k$NN-LM would require designing robust embedding and retrieval methods for low-frequency tokens and developing datastore compression techniques with fewer errors.
One approach could involve applying inverse document frequency (IDF) weights to target tokens, which would help adjust the output probabilities for informative low-frequency tokens.
Additionally, Zipfian whitening \cite{yokoi2024zipfian}, which normalizes embeddings based on token frequencies, may help mitigate frequency bias of the datastore.
We hope that our insights will contribute to exploring more effective methods with $k$NN-LM for language modeling and other natural language processing tasks in future research.

\section*{Limitations}
Our work revealed that $k$NN-LM deteriorates the prediction of low-frequency tokens, which was contrary to the prior hypothesis, and we deeply probed the reasons behind this phenomenon.
While our analyses focused on the prediction of subword tokens, extending these analyses to the level of words and phrases would be insightful.

We compared the expected values, i.e., the \emph{arithmetic} means, of $k$NN probabilities and the base LM probabilities.
Note that the evaluation metric commonly used in language modeling tasks, perplexity (PPL), corresponds to the \emph{geometric} mean of the linear interpolation of $k$NN probabilities and the base LM probabilities.
While the trends in the arithmetic means of each probability should resemble the trends in PPL, they are not completely identical.

We conducted analyses using GPT-2, which does not include Wikipedia text in its training data, to evaluate WikiText-103 in an out-of-domain scenario where $k$NN-LM is known to be most effective.
It is also known that $k$NN-LM improves language modeling performance even in in-domain settings.
Analyzing $k$NN-LM in in-domain settings remains our future work.

\section*{Acknowledgements}
This work was supported by JST SPRING Grant Number JPMJSP2140.

\bibliography{anthology,custom}

\appendix

\section{Detailed Settings}
\label{appendix:detailed_settings}
\subsection{Dataset}
\label{appendix:dataset}
\begin{figure}[t]
    \centering
    \includegraphics[width=0.85\linewidth]{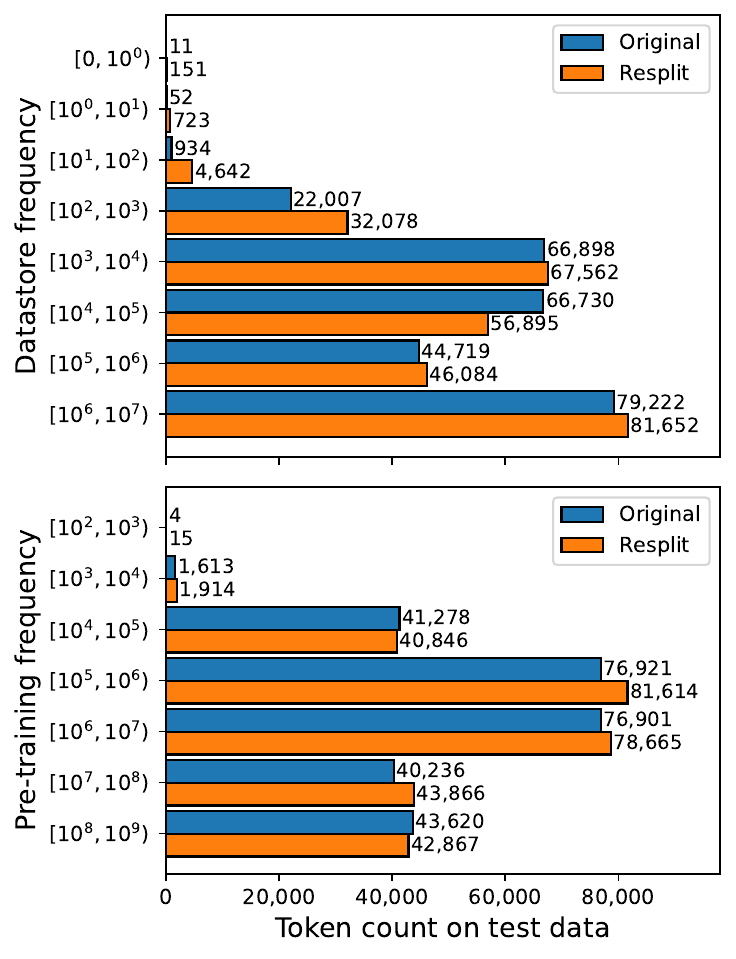}
    \caption{The number of token occurrences in the original and resplit test data of WikiText-103 for each bin of datastore frequency (upper) and pre-training frequency (lower): The resplit test data contains more low-frequency tokens than the original test data.}
    \label{fig:test_data_count}
\end{figure}

The upper plot of Figure~\ref{fig:test_data_count} presents the number of tokens in the test data of the original and resplit datasets for datastore frequency, i.e., token frequency in the corresponding training data.
The figure shows that the resplit test data contains more tokens with low datastore frequency (less than $10^3$) compared to the original data.
The lower plot of Figure~\ref{fig:test_data_count} shows the count of tokens in original/resplit test data for pre-training frequency.
The figure indicates that the resplit test data contains slightly more tokens with low pre-training frequency than the original split.
Therefore, our resplit data is more suitable for analyzing low-frequency tokens than the original data.

\subsectionmath{$k$NN-LM}
\label{appendix:detailed_knnlm}
We quantized the datastore with IVFPQ, setting the code size to 64, the number of bits to 8, and the number of centroids to 4096.
We used IndexFlatL2 as the coarse quantizer.
The index looks up 32 cluster centroids while searching for the nearest neighbors. 

\subsection{Preliminary Experiments}
\label{appendix:detailed_experiments}
In the preliminary experiments in \S\ref{sec:preliminary_experiments}, we tuned hyperparameters $k$ and $\tau$ for $k$NN-LM, selecting from $k \in \{2,4,8,16,32,64,128,256,512,1024\}$ and $\tau \in \{0.01, 0.1, 1, 10, 100, 1000\}$, minimizing PPL with each validation data of original/resplit WikiText-103.
We used $k=1024, \tau=10$ for both the original and resplit evaluation data.

\section{Further Analyses}
\begin{figure*}[t]
    \centering
    \begin{subfigure}[b]{0.45\textwidth}
        \centering
        \includegraphics[width=\textwidth]{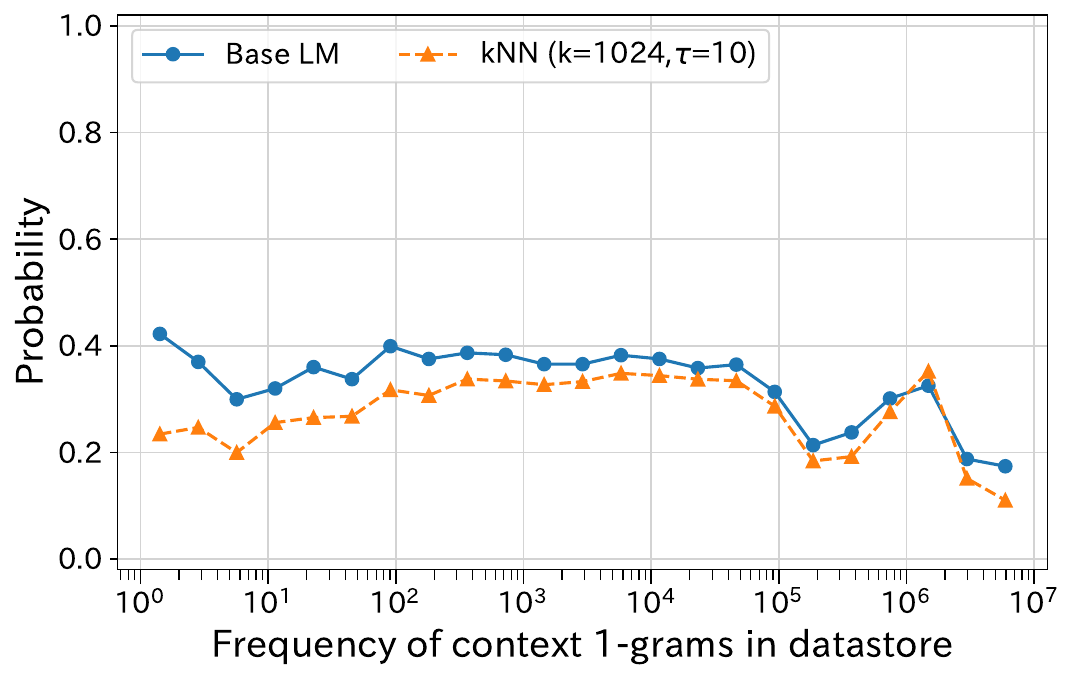}
        \caption{$n=1$}
    \end{subfigure}
    \hfill
    \begin{subfigure}[b]{0.45\textwidth}
        \centering
        \includegraphics[width=\textwidth]{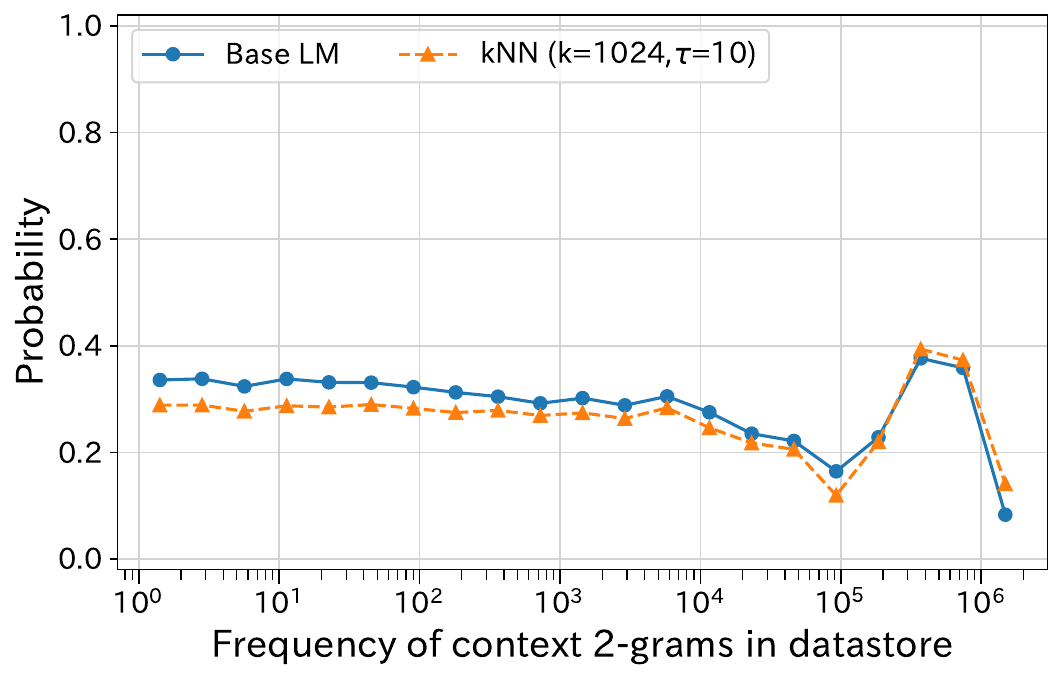}
        \caption{$n=2$}
    \end{subfigure}
    \vskip\baselineskip
    \begin{subfigure}[b]{0.45\textwidth}
        \centering
        \includegraphics[width=\textwidth]{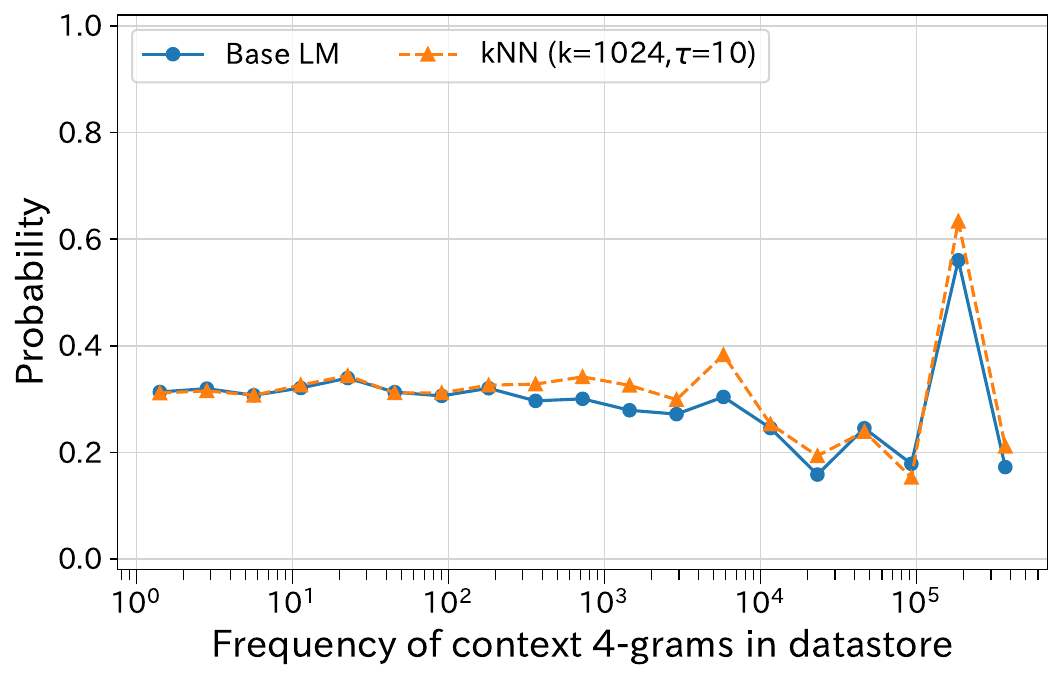}
        \caption{$n=4$}
    \end{subfigure}
    \hfill
    \begin{subfigure}[b]{0.45\textwidth}
        \centering
        \includegraphics[width=\textwidth]{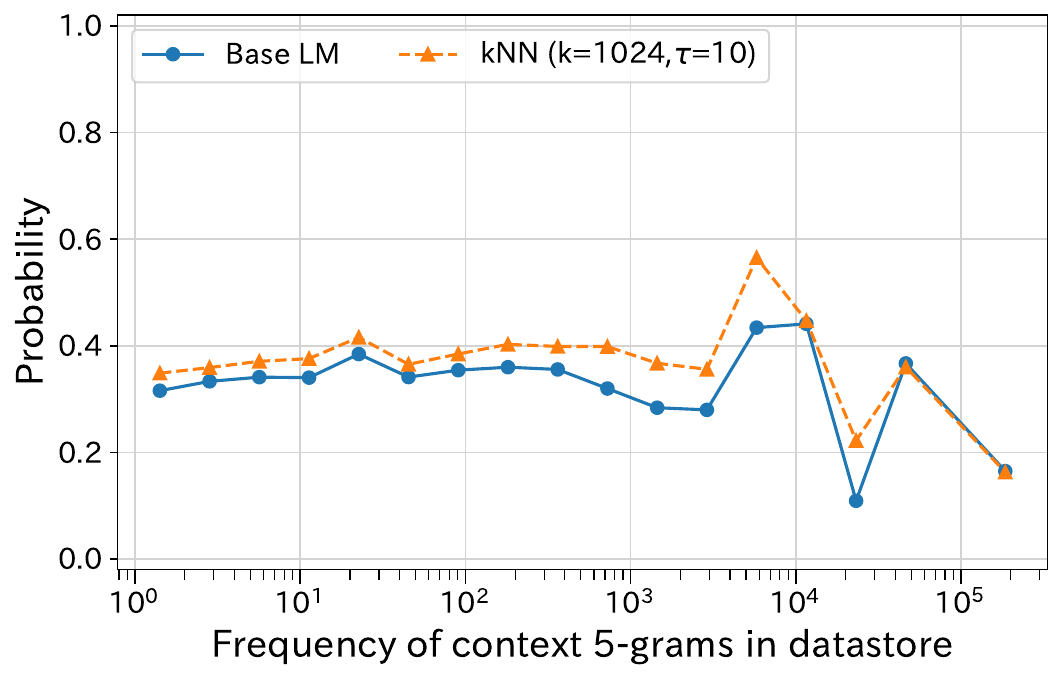}
        \caption{$n=5$}
    \end{subfigure}
    \caption{The relationship between the frequency of context $n$-grams in the datastore and the expected values of $k$NN/LM probabilities on the resplit test.}
    \label{fig:prob_vs_context_ngram}
\end{figure*}

\subsection{Predictions for long-tail contexts}
\label{appendix:context}
In \S\ref{sec:preliminary_experiments}, we showed the relationship between the frequency of context $n$-grams in the datastore and the expected values of $k$NN/LM probabilities on the resplit test for $n=3$.
Figure~\ref{fig:prob_vs_context_ngram} shows the relationship between the frequency of context $n$-grams and the probabilities for other $n$ values ($n=1, 2, 4, 5$). For these other $n$ values, we observed a similar trend to $n=3$, where no substantial improvement from $k$NN-LM was observed for low-frequency context $n$-grams.

\subsectionmath{$k$NN Probability}
\label{appendix:knn_prob}
We analyzed the relationship between token frequency, $k$NN probability, and base LM probability, revealing that $k$NN-LM worsens the prediction probabilities for low-frequency tokens in \S\ref{sec:knn_prob_analysis}.
To deeply understand, we investigate the influence of the hyperparameters, the number of neighbors $k$ and the temperature $\tau$.
To this end, we sorted the token types in the base LM's vocabulary by their datastore frequency and divided them into three equal categories: LOW, MED, and HIGH.
For each of these categories, we plot the relationship between probabilities and the hyperparameters.

\begin{figure}[t]
    \centering
    \includegraphics[width=0.95\linewidth]{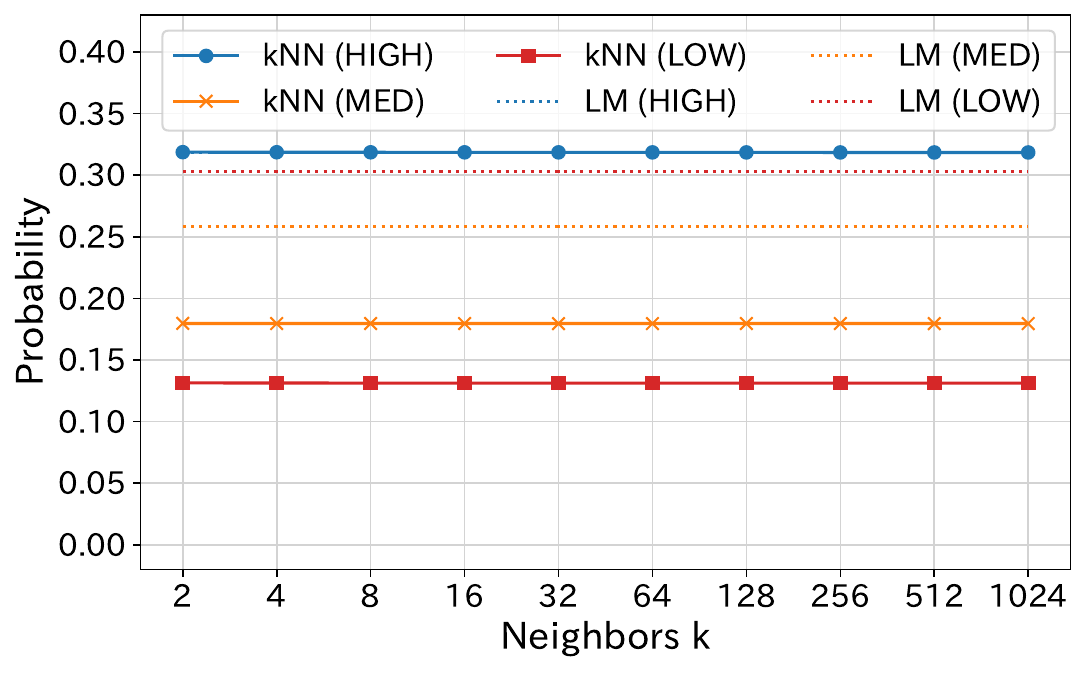}
    \caption{The relationship between $k$NN/LM probabilities and the number of neighbors $k$ across frequency categories ($\tau=1$).}
    \label{fig:knn_prob_by_k}
\end{figure}

Figure~\ref{fig:knn_prob_by_k} shows the relationship between $k$NN/LM probabilities and the number of neighbors $k$ for each token frequency category at $\tau = 1$ on the resplit test data.
The figure indicates that the $k$NN probability varies little with the number of neighbors and consistently remains lower than the LM probability for tokens in the low-frequency category, regardless of the number of neighbors $k$.

\begin{figure}[t]
    \centering
    \includegraphics[width=0.95\linewidth]{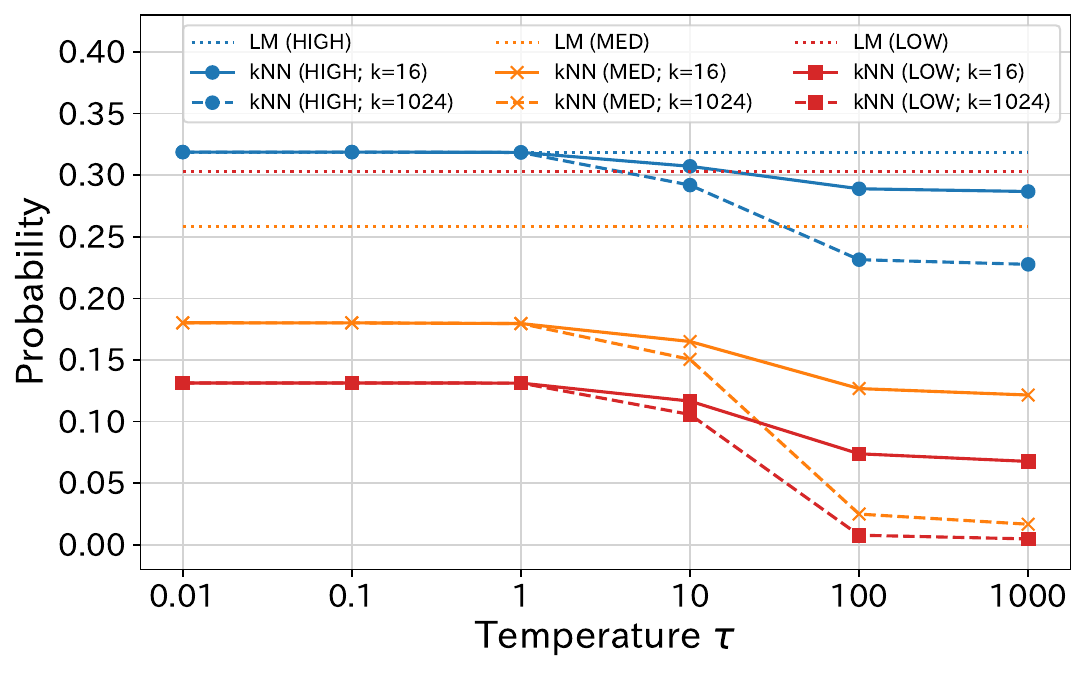}
    \caption{The relationship between $k$NN/LM probabilities and the number of temperature $\tau$ across frequency categories.}
    \label{fig:knn_prob_by_temp}
\end{figure}

Figure~\ref{fig:knn_prob_by_temp} shows the relationship between $k$NN/LM probabilities and the temperature $\tau$.
The figure reveals that increasing the temperature $\tau$ accentuates the lower $k$NN probability for the LOW frequency category.

These results indicate that the tendency of $k$NN-LM to worsen prediction probabilities for low-frequency tokens, as we revealed, cannot be resolved by adjusting the hyperparameters.

\subsectionmath{$k$NN Hit Rate}
\label{appendix:knn_hit_rate}
We demonstrated that tokens with low datastore frequency have a low $k$NN hit rate in \S\ref{sec:knn_search_analysis}.

\begin{figure}[t]
    \centering
    \includegraphics[width=0.95\linewidth]{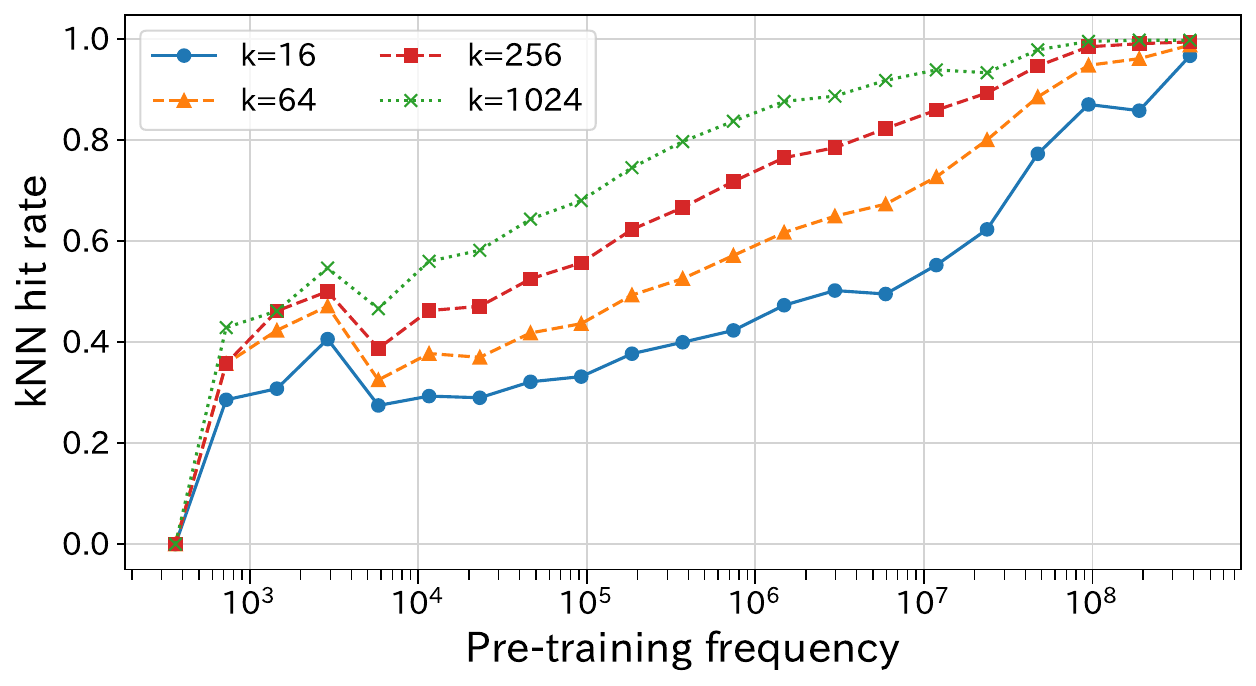}
    \caption{The relationship between pre-training frequency and the expected $k$NN hit rate on the resplit test.}
    \label{fig:knn_hit_ratio_openwebtext}
\end{figure}

Figure~\ref{fig:knn_hit_ratio_openwebtext} shows the relationship between pre-training frequency and $k$NN hit rate on the resplit test data.
Similar to the case with datastore frequency, the figure reveals that tokens with low pre-training frequency have a low $k$NN hit rate.
The majority of tokens with a frequency of approximately $10^3$ or less were not being retrieved at all, even with $k = 1024$.

\subsection{Token Distribution in Datastore}
\label{appendix:token_distribution}
We demonstrated that tokens with low datastore frequency are sparsely distributed in the datastore and that their neighbors are contaminated with different token types in \S\ref{sec:datastore_distribution_analysis}. 

\begin{figure}[t]
    \centering
    \includegraphics[width=0.95\linewidth]{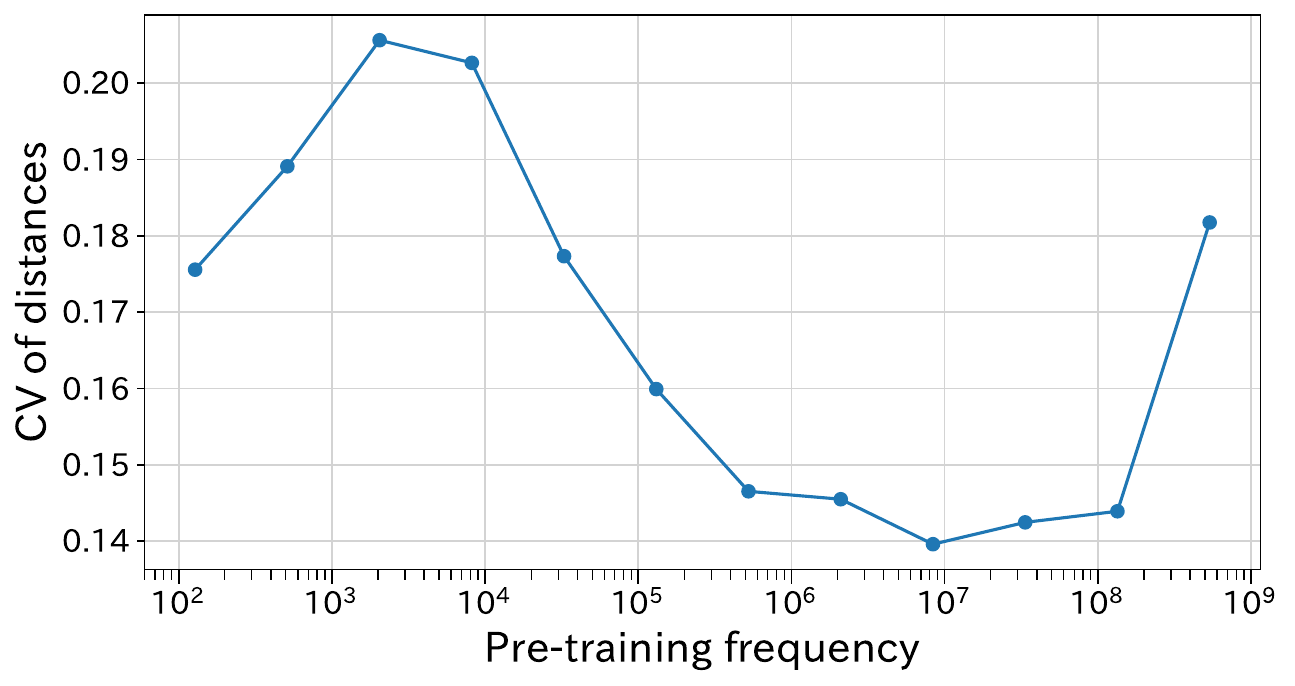}
    \caption{The relationship between pre-training frequency and the CV of the distances.}
    \label{fig:datastore_cv_openwebtext}
\end{figure}

Figure~\ref{fig:datastore_cv_openwebtext} shows the relationship between pre-training frequency and the coefficient of variation (CV) of distances on the resplit test data.
Similar to the case with datastore frequency, the figure reveals that tokens with lower pre-training frequency tend to be more sparsely distributed, except for tokens with a frequency greater than $10^8$.
Some high-frequency tokens exhibit a different trend due to their polysemy, which affects only the right tail.

\begin{figure}[t]
    \centering
    \includegraphics[width=0.95\linewidth]{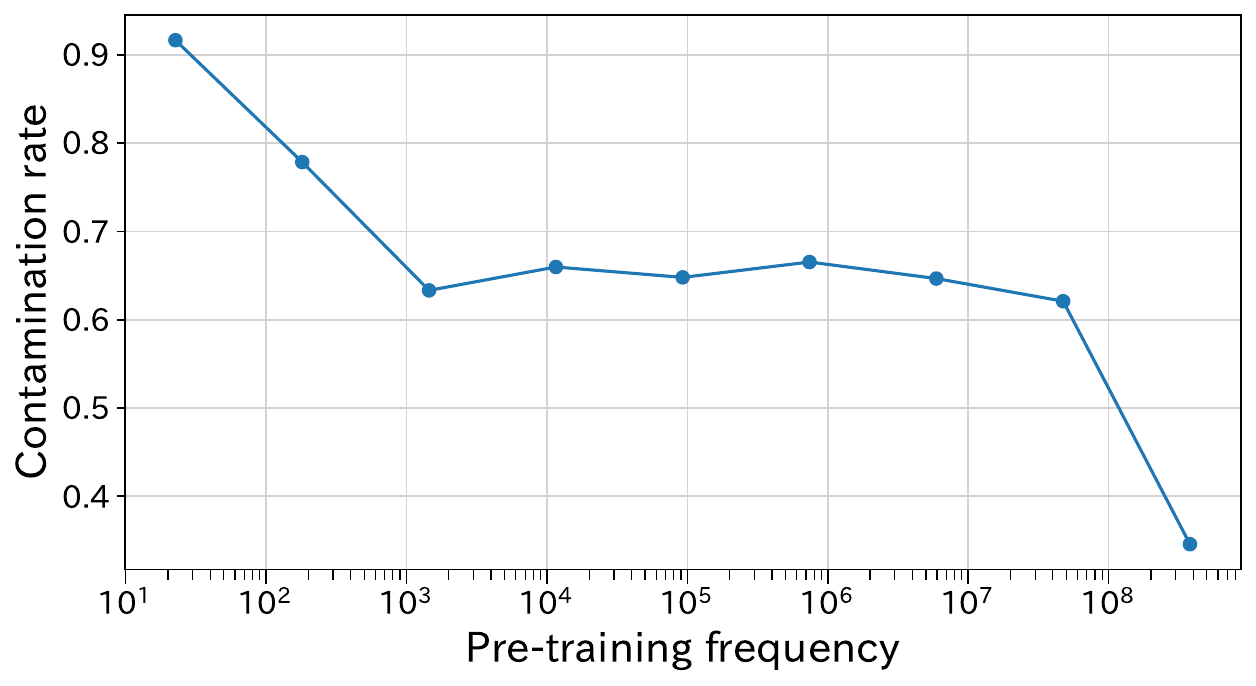}
    \caption{The relationship between token frequency of the pre-training data and the contamination rate of the token type.}
    \label{fig:datastore_contamination_openwebtext}
\end{figure}

Figure~\ref{fig:datastore_contamination_openwebtext} shows the relationship between pre-training frequency and the contamination rate on the resplit test data.
Similar to the case with datastore frequency, the figure indicates that the neighbors of tokens with lower pre-training frequency are often mixed with other token types.

\subsection{Quantization Error}
\label{appendix:quantization_error}
We demonstrated that tokens with low datastore frequency have larger approximation errors of PQ in \S\ref{sec:quantization_analysis}.

\begin{figure}[t]
    \centering
    \includegraphics[width=0.95\linewidth]{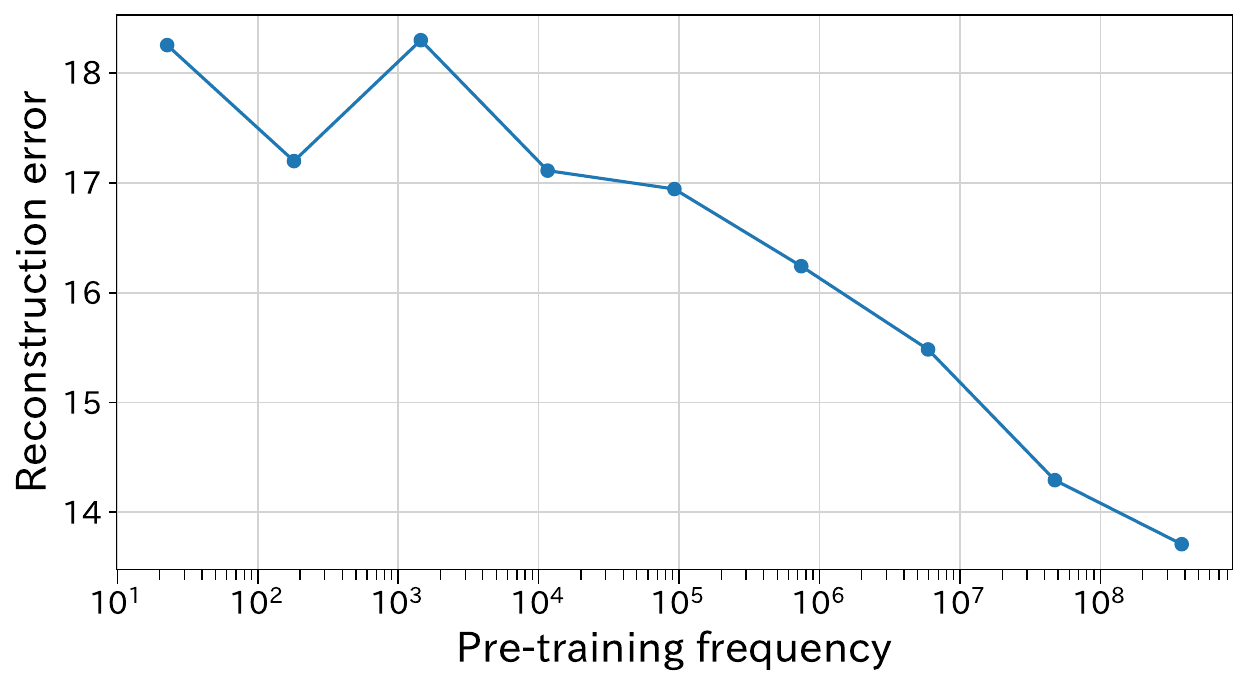}
    \caption{The relationship between token frequency of the pre-training data and the reconstruction error of the product quantization.}
    \label{fig:reconstruction_error_openwebtext}
\end{figure}

Figure~\ref{fig:reconstruction_error_openwebtext} shows the relationship between pre-training frequency and PQ reconstruction error in the resplit test data.
Similar to the case with datastore frequency, the figure reveals that tokens with lower pre-training frequency tend to have larger approximation errors.

\section{Used Data, Model, and Software}
\subsection{Data}
\begin{description}
\item[WikiText-103] created by \newcite{merity2016pointer}. License: CC BY-SA 3.0.
\item[OpenWebTextCorpus] created by \newcite{Gokaslan2019OpenWeb}. License: CC0 1.0.
\end{description}

\subsection{Model}
\begin{description}
\item[GPT2-XL] created by \newcite{radford2019language}. Download: \url{https://huggingface.co/openai-community/gpt2-xl}. License: MIT.
\end{description}

\subsection{Software}
\begin{description}
\item[\texttt{FAISS}] created by \newcite{johnson2019billion,douze2024faiss}. Download: \url{https://github.com/facebookresearch/faiss}, License: MIT.
\item[\texttt{knn-transformers}] created by \newcite{alon2022neuro}. Download: \url{https://github.com/neulab/knn-transformers}, License: MIT.
\end{description}

\end{document}